\documentclass{article}
\usepackage{spconf,amsmath,epsfig}
\usepackage{multirow}
\usepackage{threeparttable}
\usepackage{float}
\usepackage[absolute,overlay]{textpos} 

\title{Assessment of Local Climate Zone Products via Simplified Classification Rule with 3D Building Maps}

\name{Hunsoo Song, Gaia Cervini, Jinha Jung}
\address{Purdue University, West Lafayette, Indiana, USA}

\begin{document}

\begin{textblock*}{\textwidth}(2cm,1cm) 
\noindent This work is a part of the International Geoscience and Remote Sensing Symposium (IGARSS) 2023 proceedings. \\ Copyright \copyright 2023 IEEE.
\end{textblock*}

\maketitle
\begin{abstract}This study assesses the performance of a global Local Climate Zone (LCZ) product \cite{demuzere2022global}. We examined the built-type classes of LCZs in three major metropolitan areas within the U.S. A reference LCZ was constructed using a simple rule-based method based on high-resolution 3D building maps. 
Our evaluation demonstrated that the global LCZ product struggles to differentiate classes that demand precise building footprint information (Classes 6 and 9), and classes that necessitate the identification of subtle differences in building elevation (Classes 4-6).
Additionally, we identified inconsistent tendencies, where the distribution of classes skews differently across different cities, suggesting the presence of a data distribution shift problem in the machine learning-based LCZ classifier. Our findings shed light on the uncertainties in global LCZ maps, help identify the LCZ classes that are the most challenging to distinguish, and offer insight into future plans for LCZ development and validation.

\end{abstract}

\section{Introduction}


\begin{figure*}[t]
	\centering
	\includegraphics[width=0.95\textwidth]{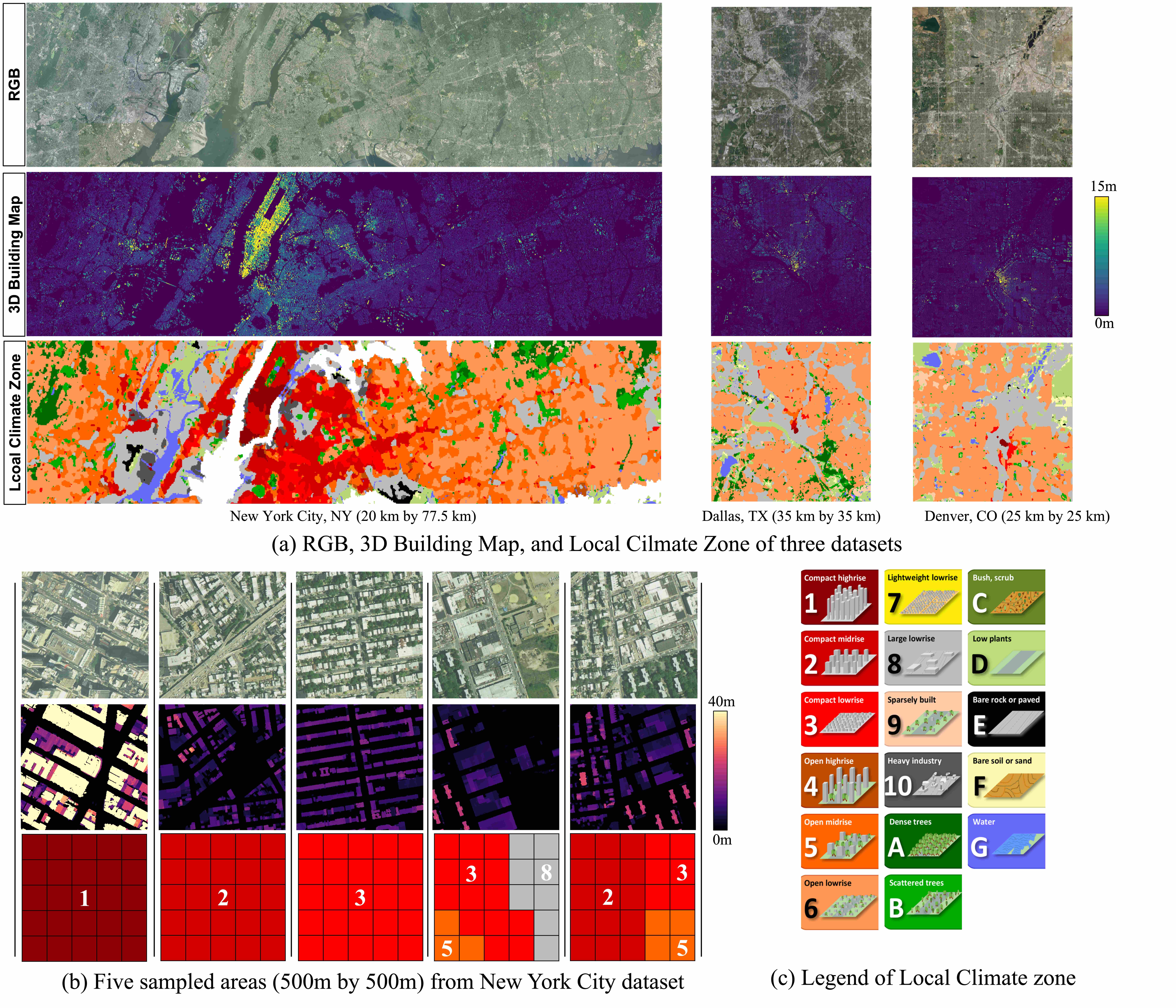}
        \vspace{-0.2in}
	\caption{RGB, 3D Building Map, and Local Climate Zone from experimental datasets ((c) was adapted from \cite{demuzere2020combining})}
    \vspace{-0.1in}
\end{figure*}

Urban geospatial data have been widely used to investigate the current status of cities and to develop urban design strategies for sustainable urbanization. Many studies have developed a variety of map classification schemes that can abstract an urban landscape into a form that can better illustrate its impacts on urban environments. The local climate zone (LCZ) is one successful classification scheme that classifies land cover into 17 classes based on its physical properties related to urban climate \cite{stewart2012local}. In particular, the LCZ classes are closely associated with the urban 3D structure, unlike the typical land cover classification system \cite{homer2020conterminous,zhang2022urbanwatch, brown2022dynamic}.

In 2022, a global LCZ map with a 100m resolution was released \cite{demuzere2022global}. The LCZ classification was performed using a supervised random forest classifier that used a large amount of labeled training samples and diverse earth observation inputs. The earth observation inputs include earth observation satellite images and other sources that provide textual and height information \cite{demuzere2020combining, demuzere2021lcz}. While the LCZ classifier integrated 3D urban structure information into its classification process, we hypothesize that some errors or biases may be present, particularly when characterizing 3D urban features. This is because the height input for the LCZ classifier is largely based on approximate height information with coarse resolution \cite{demuzere2022global}.

This research assesses the recent global LCZ product based on a simplified LCZ classification rule that employs high-resolution 3D building maps generated from airborne LiDAR data.

\section{Datasets and Methods}

We conducted our evaluation using three geographically diverse and distantly located metropolitan areas: New York City, NY; Dallas, TX; and Denver, CO. In particular, our assessment primarily focused on built-type classes whose classifications are notably affected by 3D building information. Specifically, LCZ Classes 1-6, 8, and 9 were examined, due to the scarcity of Classes 7 and 10 in the datasets.

For creating a reference, we ``re-classified'' three metropolitan areas using a simple classification rule. First, we created large-scale 3D building maps at 1 m resolution for these areas (Fig. 1). These maps were generated utilizing an open-source building mapping algorithm \cite{song2022towards, song2022challenges} applied to LiDAR data from the U.S. Geological Survey's 3D Elevation Program \cite{stoker2022accuracy}. We then matched a 100m x 100m 3D building map tile with its corresponding 100m resolution LCZ map. Utilizing the 3D building map, we calculated the Building Surface Fraction (BSF) and the Height of Roughness Elements (HRE) \cite{stewart2012local}. These measures enabled the classification of Classes 1-6 and Class 9. Subsequently, we ``post''-classified any instances deemed either Class 1-6 or Class 9 into Class 8, if they met the criteria based on Sky View Factor (SVF) and Pervious Surface Fraction (PSF). SVF was calculated from the digital surface model, and PSF was computed using NDVI from the National Agriculture Imagery Program (NAIP) and a surface water map derived from \cite{song2023water}.

\begin{table}[h]
\centering
\begin{threeparttable}
\vspace{-0.05in}
\caption{Simplified LCZ classification rules}
\label{tab:classification_rules}
\begin{tabular}{cccc}
\hline
\textbf{Class} & \multicolumn{3}{c}{\textbf{Classification rule}} \\
\hline
1 & BSF $>$ 0.4 & \& & HRE $>$ 25 \\
2 & BSF $>$ 0.4 & \& & $10 \leq$ HRE $\leq$ 25 \\
3 & BSF $>$ 0.4 & \& & $3 \leq$ HRE $\leq$ 10 \\
4 & $0.2 \leq$ BSF $\leq$ 0.4 & \& & HRE $>$ 25 \\
5 & $0.2 \leq$ BSF $\leq$ 0.4 & \& & $10 \leq$ HRE $\leq$ 25 \\
6 & $0.15 \leq$ BSF $\leq$ 0.25 & \& & $3 \leq$ HRE $\leq$ 10 \\
8 & SVF $>$ 0.7 & \& & PSF $<$ 0.2 \\
9 & $0.05 \leq$ BSF $\leq$ 0.15 & \& & $3 \leq$ HRE $\leq$ 10 \\
Others & \multicolumn{3}{c}{\emph{N/A}} \\
\hline
\end{tabular}
\begin{tablenotes}
\footnotesize	
\item[1] BSF: Building Surface Fraction, HRE: Height of Roughness Elements, SVF: Sky View Factor, PSF: Pervious Surface Fraction.
\vspace{-0.2in}
\end{tablenotes}
\end{threeparttable}
\end{table}

\begin{table*}[h]
\vspace{-0.05in}
\footnotesize
\renewcommand{\arraystretch}{1}
\caption{Comparison of the number of pixels between original and reclassified LCZ maps by class}
\label{table_example}
\centering
\begin{tabular}{c c ccc ccc cccc}
\hline
 & \multirow{2}{*}{Dataset} & \multicolumn{10}{c}{LCZ Class} \\
\cline{3-12}
 &  & 1 & 2 & 3 & 4 & 5 & 6 & 7 & 8 & 9 & 10\\
\hline
\multirow{3}{*}{Original LCZ \cite{demuzere2022global}} & New York City & 2470 & 22852 & 9377 & 2899 & 42994 & 68018 & 0 & 29103 & 223 & 3897 \\
 & Dallas        & 144 & 248 & 773 & 10 & 1886 & 82174 & 0 & 38008 & 2670 & 0\\
 & Denver        & 163 & 32 & 985 & 23 & 696 & 46761 & 0 & 19273 & 2784 & 0\\
 \hline
\multirow{3}{*}{Reclassified LCZ} & New York City & 1672 & 5658 & 7439 & 1307 & 7511 & 48034 & 0 & 7747 & 46238 & 0\\
 & Dallas & 171 & 874 & 1577 & 195 & 1628 & 32125 & 0 & 8456 & 32467 & 0\\
 & Denver & 98 & 381 & 388 & 107 & 951 & 17315 & 0
& 6587 & 25347 & 0\\
\hline
\end{tabular}
\end{table*}

It is important to note that our simplified LCZ classification scheme is mutually exclusive, though not collectively exhaustive. The proposed rule does not account for all conditions of LCZ classification and cannot distinguish all 17 LCZ classes. However, the classification rule is explicitly defined and anchored to the standard LCZ classification scheme \cite{stewart2012local}. This clarity renders the classification results a reliable reference LCZ for evaluation. Particularly, due to the high accuracy of the airborne LiDAR-based 3D building map, this reclassified LCZ can serve as a highly reliable reference for evaluation, especially in terms of 3D elevation accuracy.

\section{Experimental results}

\begin{figure*}[h!]
        \vspace{-0.05in}
	\centering
	\includegraphics[width=6.25in]{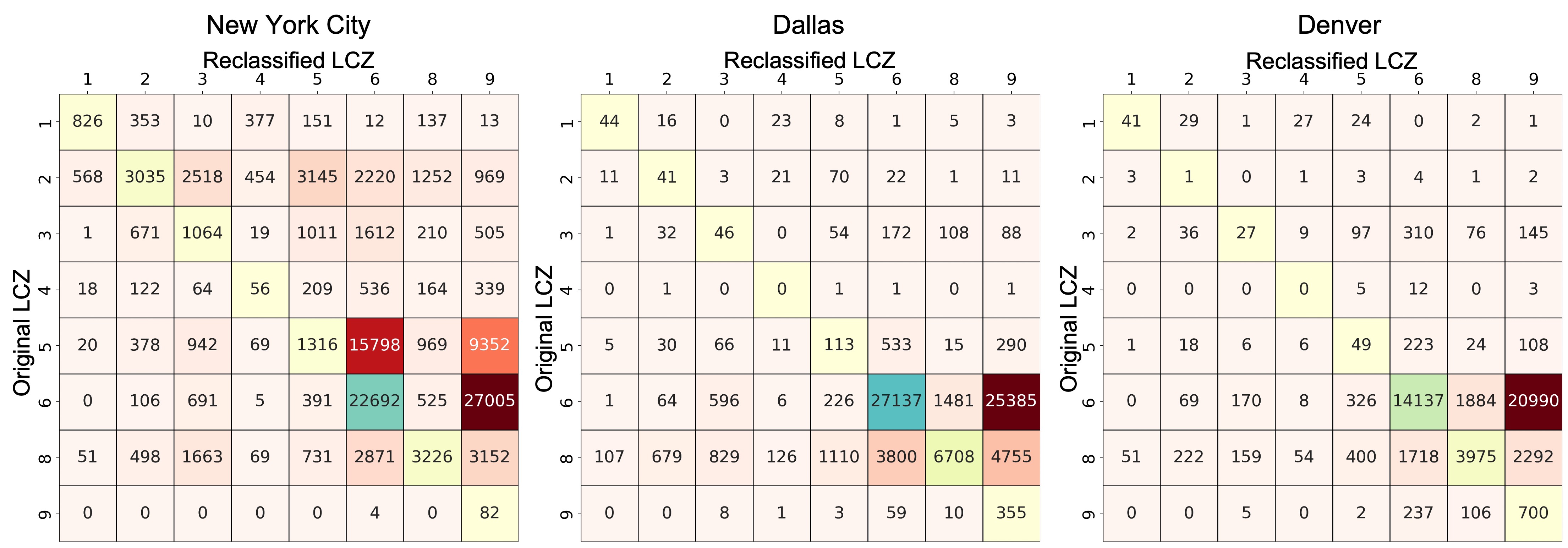}
	\caption{Confusion matrices comparing the original LCZ and the reclassified LCZ for three different cities}
\end{figure*}

\begin{figure*}[h!]
        \vspace{-0.05in}
	\centering
	\includegraphics[width=6.25in]{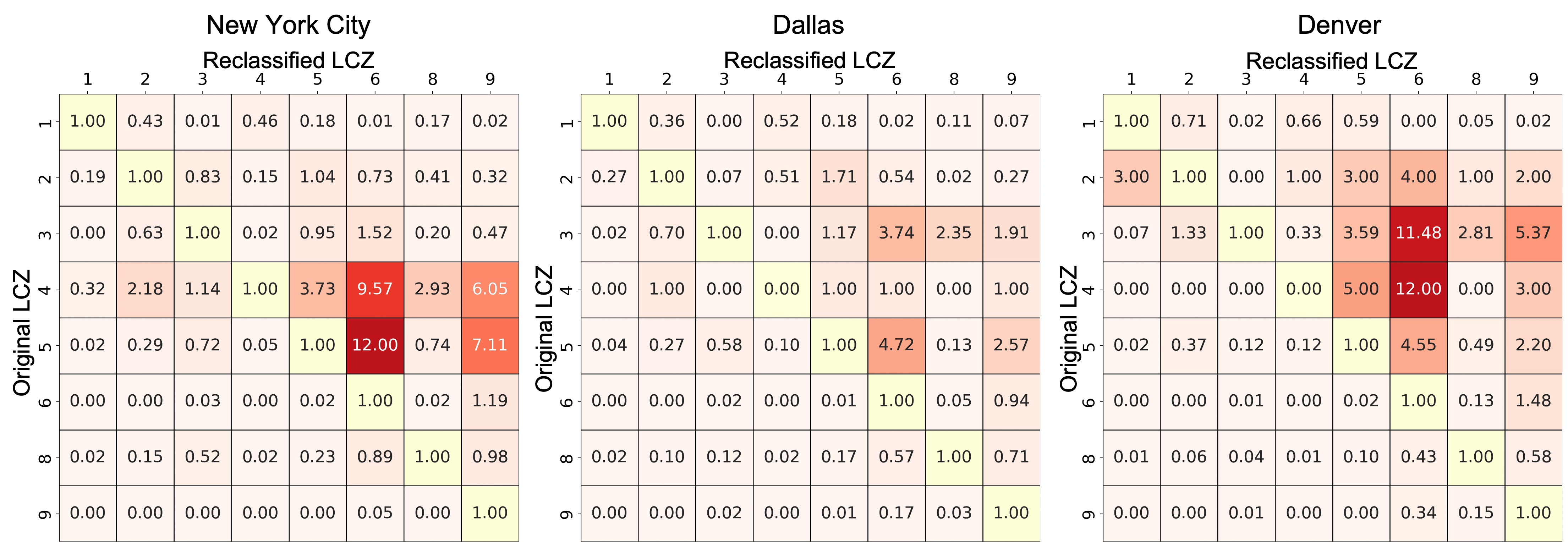}
	\caption{Row-wise normalized confusion matrices comparing the original LCZ and the reclassified LCZ for three different cities}
\end{figure*}

Table 2 and Fig. 2. show the comparative distribution of LCZ classes and confusion matrix, respectively, between the LCZ map of the global product (``original LCZ'') and the LCZ map resulting from the simplified rule (``reclassified LCZ''). For Fig. 3., we normalize the confusion matrices' (Fig. 2.) each row by its diagonal element to identify where the global LCZ product is likely to get confused. If any off-diagonal element exceeds 1, it indicates the global LCZ product is significantly perplexed by the class of the off-diagonal element.

As depicted in Figures 2-3, ``heated'' areas demonstrate several common tendencies. In Classes 1 and 8, the two different LCZs show a relative alignment. However, there is a notable lack of agreement in Classes 3, 4, and 5. Furthermore, as demonstrated in Table 2, the count of Class 9 is significantly underestimated across all cities, mainly due to confusion with Class 6, which is predominant in the three datasets. If we assume that the primary distinction between Classes 6 and 9 lies in the Building Surface Fraction (BSF), it appears that the classifier for the global LCZ product struggles with identifying detailed building information and tends to be heavily biased toward Class 6.

In addition, Classes 4 and 5 of the global LCZ product are often reclassified as either 5 or 6 in the reference LCZ. Given that Classes 4, 5, and 6 primarily vary based on the overall elevation of buildings, it suggests that the global LCZ product struggles in differentiating subtle elevation differences in ``open urban areas (Classes 4, 5, and 6)''. Considering that the global LCZ product is primarily based on spectral characters and coarse elevation products, the classification challenge among Classes 4, 5, and 6 is not surprising.

Interestingly, while global LCZ products across different cities face similar classification challenges, the areas of major disagreement vary significantly among the cities, suggesting the presence of a data distribution shift. For instance, as depicted in Figure 3, the case of New York City shows the highest disagreement rates in Classes 4 and 5, whereas Dallas and Denver have the most pronounced disagreements in Class 3. The reference LCZ is created based on a rule that applies uniformly across all cities, with Classes 3-5 determined exclusively by the 3D building map. As such, we can anticipate that the reference LCZ will be considerably less affected by data distribution shifts. Given this robustness, the observed discrepancies across cities indicate that the classifier for the global LCZ product might have experienced data distribution shifts. This suggestion is further supported by the imbalanced ratios across the three datasets, as shown in Table 2. These inconsistencies could potentially stem from variations in spectral characteristics or discrepancies in the accuracy of GIS products used for mapping across different cities.

\section{Summary and Future Remarks}

The global LCZ product \cite{demuzere2022global} was assessed based on high-resolution 3D building maps, focusing on built-type classes of three metropolitan cities in the U.S. A reference LCZ map was generated using a simple yet robust classification rule with 3D building maps. We found the global LCZ product tends to underestimate Class 9 while overestimating Classes 6 and 8. Also, the product exhibits limited capability in discerning subtle elevation differences among Classes 4-6. Moreover, noticeable inter-city biases in the distribution of these classes were observed.

The significance of machine learning-based LCZ mapping is unquestionable, especially considering the scarcity of high-resolution 3D elevation products. Nonetheless, in light of the rapidly changing landscape of available geospatial data, such as airborne LiDAR data, it may be necessary to make amendments to the LCZ classification method and scheme. Also, these changes should be accompanied by thorough validation, to ensure consistent outcomes across large areas and to reliably extract knowledge from multi-city studies. Although this might compromise the role of LCZ as a tool for characterizing climate-land interactions, a mapping method based on more definitive features—such as the number of buildings, rather than the Aspect Ratio, which lacks a standardized method—could enhance consistency across extensive areas and foster more generalizable results. 

Our current simple classification rule has limitations in capturing non-linear, complex features, which are essential for comprehensive LCZ classification. Nonetheless, our evaluation results—obtained through a methodology that ensures high consistency across different cities—effectively reveal the existing challenges in LCZ classifications. We anticipate that our evaluation methods and results will contribute to the development of a reliable validation tool and offer valuable insights for the future enhancement of LCZ products.
\bibliographystyle{IEEEbib}
\bibliography{references}

\end{document}